\documentclass{article}

     \usepackage[nonatbib,preprint]{neurips_2019}

\usepackage[utf8]{inputenc} 
\usepackage[T1]{fontenc}    
\usepackage{url}            
\usepackage{booktabs}       
\usepackage{nicefrac}       
\usepackage{microtype}      

\usepackage{amsmath,amssymb,amsfonts}
\usepackage{graphicx, import}
\usepackage{textcomp}
\usepackage[center]{subfigure}
\usepackage{multirow}
\usepackage{bm}
\usepackage{siunitx}

\usepackage{todonotes}

\usepackage{algorithm}
\usepackage[noend]{algpseudocode}

\usepackage[pdfstartview=FitH,
             plainpages=false,
             bookmarksopen=true,
             bookmarksnumbered=true,
             colorlinks=true,
             bookmarkstype=toc]{hyperref} 

\usepackage{xcolor} 

\usepackage{upgreek}

\renewcommand{\v}[1]{{\boldsymbol{\mathbf{#1}}}}

\newcommand\latent{\v{z}}
\newcommand\obs{\v{x}}
\newcommand\action{\v{a}}
\newcommand\muvec{\v{\mu}}

\makeatletter
\newcommand{\mquote}[1]{{\mathpalette\mqu@te{#1}}}

\newcommand{\mqu@te}[2]{%
  \sbox0{$\m@th#1\text{``}$}%
  \sbox2{$\m@th#1\text{''}$}%
  \sbox4{$\m@th#1#2$}%
  \ifdim\ht4>\dimexpr\ht0+1pt\relax
    \raisebox{\dimexpr\ht4-\height}{\box0}%
    #2%
    \raisebox{\dimexpr\ht4-\height}{\box2}%
  \else
    \box0 #2\box2
  \fi
}
\makeatother

\graphicspath{{graphics/},{logs/}}

\title{End-to-End Pixel-Based Deep Active Inference for Body Perception and Action}

\author{
Cansu Sancaktar\\ Technical University of Munich\\ Germany\\ \texttt{cansu.sancaktar@tum.de}
\And
Marcel A. J. van Gerven\\Donders Institute for Brain, Cognition and Behaviour\\
Radboud University\\
Netherlands\\ \texttt{m.vangerven@donders.ru.nl}
\And
Pablo Lanillos\thanks{This work has been supported by SELFCEPTION project EU Horizon 2020 Programme, grant nr. 741941.}\\
 Donders Institute for Brain, Cognition and Behaviour\\ Radboud University\\Netherlands\\ \texttt{p.lanillos@donders.ru.nl}
}

\begin{document}

\maketitle

\begin{abstract}
We present a pixel-based deep active inference algorithm (PixelAI) inspired by human body perception and action. Our algorithm combines the free energy principle from neuroscience, rooted in variational inference, with deep convolutional decoders to scale the algorithm to directly deal with raw visual input and provide online adaptive inference. Our approach is validated by studying body perception and action in a simulated and a real Nao robot. Results show that our approach allows the robot to perform 1) dynamical body estimation of its arm using only monocular camera images and 2) autonomous reaching to ``imagined" arm poses in visual space. This suggests that robot and human body perception and action can be efficiently solved by viewing both as an active inference problem guided by ongoing sensory input.
\end{abstract}


\section{Introduction}
\label{sec:intro}
Learning and adaptation are two core characteristics that allow humans to perform flexible whole-body dynamic estimation and robust actions in the presence of uncertainty~\cite{wolpert2011principles}. We hypothesize that the human brain acquires a representation (model) of the body, already starting at the earliest stages of life, by learning a mapping between tactile, proprioceptive and visual cues~\cite{yamada2016embodied}. Furthermore, this mapping is flexible, as experiments have demonstrated that body perception and action can be altered in less than one minute just by synchronous visuotactile stimulation~\cite{botvinick1998rubber, hinz2018drifting}. This supports the view that unsupervised learning mechanisms must be enhanced by online supervised adaptation~\cite{doya1999computations}. On the contrary, robots usually use a fixed-rigid body model where the arm end-effector is defined as a pose, i.e., a 3D point in the space and orientation. Hence, any error in the model or change in the conditions will result in failure.

Several solutions have been proposed to overcome this problem, usually separated in perception and control approaches. For instance, by working in  visual space (e.g. visual servoing~\cite{hutchinson1996tutorial}) we can exploit a set of invariant visual keypoints to provide control that incorporates real-world errors. Bayesian sensory fusion in combination with model-based fitting allows adaptation to sensory noise and model errors~\cite{cifuentes2016probabilistic} and model-based active inference provides online adaptation in both action and perception~\cite{oliver2019active}. Finally, learning approaches have shown that the difference between the model and reality can be overcome by optimizing the body parameters or by explicitly learning the policy for a task, e.g., through imitation learning or reinforcement learning (RL). Recently, model-free approaches, particularly deep RL, have demonstrated the potential for directly using raw images as an input for learning visual control policies~\cite{levine2016end}.

\begin{figure}[t!]
	\centering
	\includegraphics[width=0.90\linewidth]{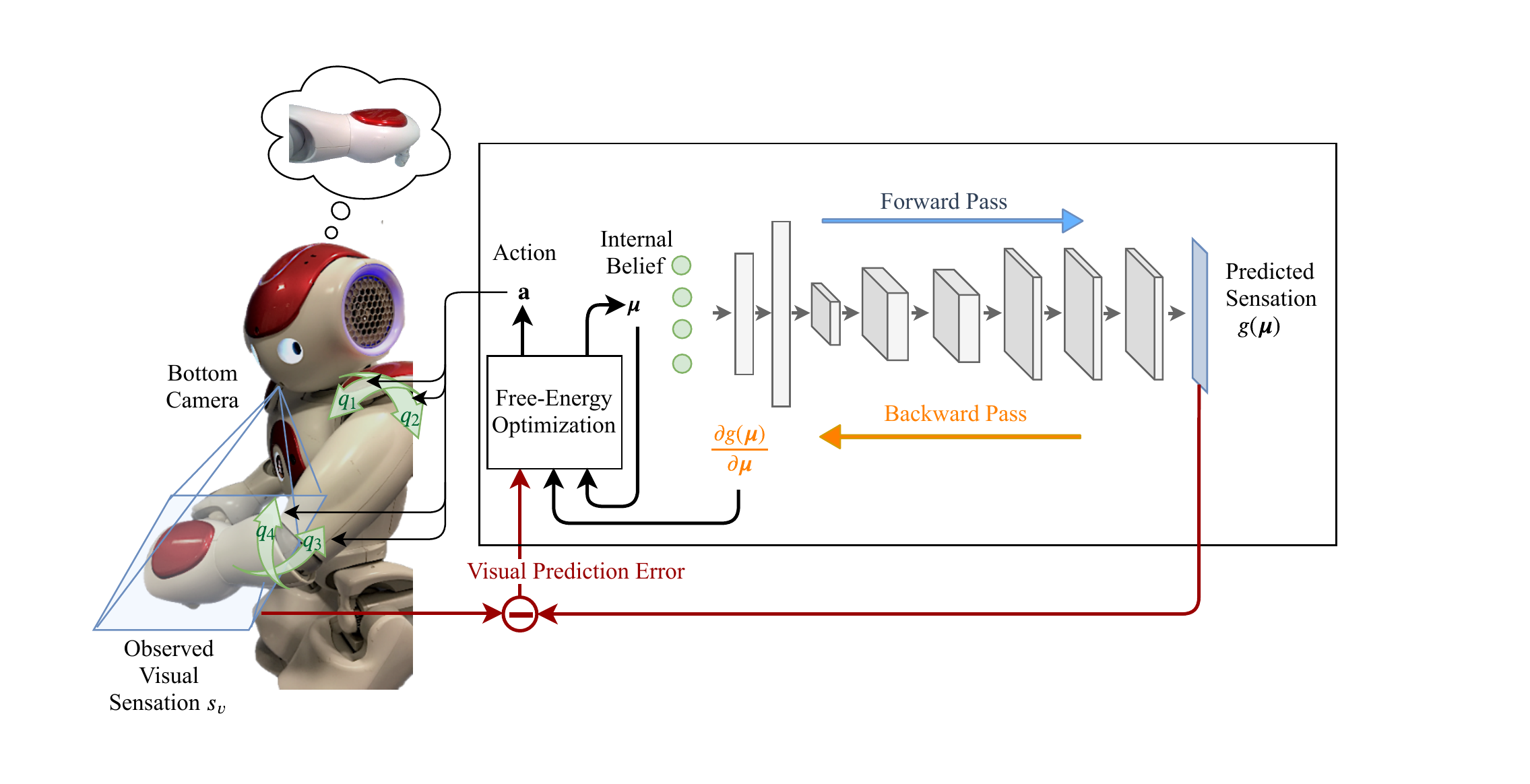}
	\caption{Pixel-based deep active inference (PixelAI). The robot infers its body (e.g., joint angles) by minimizing the visual prediction error, i.e. discrepancy between the camera sensor value $\obs_v$ and the expected sensation $\v{g}(\v{\mu})$ computed using a convolutional decoder. The error signal is used to update the internal belief to match the observed sensation (perceptual inference) and to generate an action $\v{a}$ to reduce the discrepancy between the observed and predicted sensations. Both are computed by minimizing the variational free-energy bound.
	}
	\label{img:algorithm}
\end{figure}

In this work, we introduce {PixelAI}, a pixel-based deep active inference algorithm, depicted in Fig.~\ref{img:algorithm}, which combines all of these characteristics. That is, it directly operates on visual input, provides adaptation, model-free learning and, moreover, it unifies perception and action into a single variational inference formulation.

We motivate our proposed approach using body perception and action (though it can easily be generalized to other active perception tasks). Our approach is grounded in human embodied research through the maxim \textit{first the body then the policy}~\cite{lanillos2016yielding}. That is, an agent must first learn to perceive its body (the cross-modal sensorimotor relations~\cite{diez2019sensorimotor,man2019homeostasis}) and afterwards should use this information to adapt to online changes when interacting in the world. 

Technically, our method combines the free energy principle~\cite{friston2010unified}, which updates an internal model of the body through perception and action, with deep convolutional decoders, that map internal beliefs to expected sensations.
More concretely, the agent learns a generative model of the body to aid in the construction of expected sensations from incoming partial multimodal information. This information, instead of being directly encoded, is processed by means of the error between the predicted outcome and the current input. In this sense, we formalize body perception and action as a consequence of surprise minimization via prediction errors~\cite{rao1999predictive,friston2010unified}.


Following this approach, the robot should learn a latent representation (“state”) of the body and the relation between its state and the expected sensations (“predictions”). These predictions will be compared with the observed sensory data, generating an error signal that can be propagated to refine the belief that the robot has about its body state. Compensatory actions would follow a similar principle and will be exerted to better correspond to the prediction made by the models learnt, giving the robot the capacity to actively adjust to online changes in the body and the environment (see Fig.~\ref{img:algorithm}). This also offers a natural way to realize a plan by setting an imagined goal in sensory space~\cite{nair2018visual}. For example, when working in visual space we can provide an image as a goal. By means of optimizing the free-energy bound the agent then executes those actions that minimize the discrepancy between expected sensations (our goal) and observed sensations. 
Our approach was validated by studying body perception and action in a simulated and a real Nao robot.
Results show that our approach allows the robot to perform 1) dynamical body estimation of its arm using only raw monocular camera images and 2) autonomous reaching to arm poses provided by a goal image.


\section{Background}
\label{sec:background}

\subsection{The free energy principle}
We model body perception as inferring the body state $\latent$ based on the available sensory data $\obs$. Given a sensation $\obs$, the goal is to find $\obs$ such that the posterior $p(\latent | \obs)
=p(\obs | \latent) p(\latent)/p(\obs)$. However, computing the marginal likelihood $p(\obs)$ requires an integration over all possible body states. That is, $p(\obs) = \int_{\latent}p(\obs | \latent) p(\latent) d\latent$, which becomes intractable for large state spaces. The free-energy \cite{hinton1994autoencoders}, largely exploited in machine learning~\cite{Wainwright2007} and neuroscience~\cite{friston2007variational}, circumvents this problem by introducing a reference distribution (also called recognition density) $q(\latent)$ with known tractable from. The goal of the minimization problem hence becomes finding the reference distribution $q(\latent)$ that best approximates the posterior $p(\latent | \obs)$.

For tractability purposes, this approximation is calculated by optimizing the negative variational free energy $F$, also referred to as the evidence lower bound (ELBO). $F$ can be defined as the Kullback-Leibler divergence $D_{\textrm{KL}}$ minus the negative log-evidence or sensory surprise $-\ln p(\obs)$:
\begin{align}
\label{eq:KL}
    F &= D_{\textrm{KL}}(q(\latent) \Vert p(\latent|\obs)) - \ln p(\obs) = 
     \int_\latent q(\latent) \ln\frac{q(\latent)}{p(\latent | \obs)} d\latent - \ln p(\obs) \,,
\end{align}
which, due to the non-negativity properties of $D_{\textrm{KL}}$, is an upper bound on surprise.
Alternatively, we can use the identity $\ln p(\obs) = \int_\latent q(\latent) \ln p(\obs) d\latent$ to include the second term into the integral and write Equation~(\ref{eq:KL}) as
\begin{align}
      F = \int_\latent q(\latent) \ln\frac{q(\latent)}{p(\obs, \latent)} d\latent = -\int_\latent q(\latent) \ln p(\obs,\latent) d\latent + \int_{\latent} q(\latent) \ln q(\latent) d\latent \,.
      \label{eq:fep}
\end{align}

According to the free energy principle \cite{friston2010unified} both perception and action optimize the free energy and hence minimize surprise:
 \begin{enumerate}
 \item {\em Perceptual inference}: The agent updates its internal belief by approximating the conditional density (inference), maximizing the likelihood of the observed sensation:
 \begin{align}
     \latent = \underset{\latent}{\mathrm{arg\,min}}\,  F(\latent,\obs) \,.
 \end{align}
 \item {\em Active inference}: The agent generates an action $\action$ that results in a new sensory state $\obs(\action)$ that is consistent with the current internal representation:
  \begin{align}
     \action =  \underset{\action}{\mathrm{arg\,min}}\, F(\latent,\obs(\action)) \,.
 \end{align}
 \end{enumerate}


Under the Laplace approximation, the variational density can take the form of a Gaussian $q(\latent) = \mathcal{N}(\v{\mu}, \v{\v{\Sigma}})$, where $\v{\mu}$ is the conditional mode and $\v{\Sigma}$ is the covariance of the parameters. By incorporating this reference distribution in Equation~(\ref{eq:fep}), the free-energy can be approximated as\footnote{See \cite{friston2007variational} for full derivation.},
\begin{equation}
\label{eq:F_asL}
F = - \ln p(\obs,\muvec) - \frac{1}{2}\left( \ln |\v{\v{\Sigma}}| + n \ln 2\pi \right),
\end{equation}
where the first term is the joint density of the observed and the latent variables with $\muvec$ an $n$-dimensional state vector.



\section{Pixel-based Deep Active Inference}
\label{sec:pixelAI}

Our proposed PixelAI approach combines free energy optimization with deep learning to directly work with images as visual input. The optimization provides adaptation and the neural network incorporates learning of high-dimensional input. We frame and experimentally validate the proposed algorithm in body perception and action in robots. Figure~\ref{img:algorithm} visually describes PixelAI. The agent first learns the approximate generative forward models of the body, implemented here as convolutional decoders.
While interacting, the expected sensation (predicted by the decoder) is compared with the real visual input and the prediction error is used to 1) update the belief and 2) generate actions. This is performed by means of optimizing the variational free-energy bound.


\subsection{Active inference model}

We formalize body perception as inferring the unobserved body state $\v{\mu}$, e.g., the estimation of the robot joint angles like shoulder pitch and roll. We define the robot internal belief as an $n$-dimensional vector: $\v{\mu}^{[d]} \in \mathbb{R}^{n}$ for each temporal order $d$. For instance, for first-order (velocity) generalized coordinates the belief is: $\muvec = \{\muvec^{[0]},\muvec^{[1]}\}$. The observed variables $\obs$ are the visual sensory input $\obs_v$ and the external causal variables $\v{\rho}$: $\obs=\{\obs_v,\v{\rho}\}$ For instance, the robot has access to visual information $\obs_v$ (an image of size $w \times h$) and proprioception information (the joint encoder values $\v{q}$). The causal variables $\v{\rho}$ are independent variables that produce effects in the world. Finally, let us define two generative models that describe the system. The sensory forward model $\v{g}$, which is the predictor that computes the sensory outcome $\obs_v$ given the internal state $\v{\mu}$ and the body internal state dynamics $\v{f}$. Both functions can be considered as the approximations that the agent has about the reality:
\begin{align}
&\obs_v = \v{g}(\v{\mu}) + \v{w}_v\\
&\v{\mu}^{[1]} = \v{f}(\v{\mu},\v{\rho}) + \v{w}_{\mu}
\end{align}
where $\v{w}_v, \v{w}_\mu$ are both processes noise and are assumed to be drawn from a multivariate normal distribution with zero mean and covariance $\Sigma_v$ and $\Sigma_\mu$ respectively.

In order to compute the variational free-energy under the Laplace approximation from Equation~(\ref{eq:F_asL}) we need the joint density. Assuming independence of the observed variables: 
\begin{equation}
\ln p(\obs,\v{\mu}) = \ln p(\obs_v,\v{\rho}, \v{\mu}) = \ln p(\obs_v|\v{\mu}) + \ln p(\v{\mu}^{[1]}|\v{\mu}^{[0]}, \v{\rho})
\label{eq:neg-free-energy-term}
\end{equation}
where $p(\obs_v|\v{\mu})$ is the likelihood of having a visual sensation given the internal state and $p(\v{\mu}^{[1]}|\v{\mu}^{[0]}, \v{\rho})$ is the transition dynamics of the latent variables (body state).

Body perception is then instantiated as computing the body state that minimizes the variational free-energy. This can be performed through gradient optimization $\partial F/\partial\muvec$. Since the temporal difference $\v{\mu}^{[0]}_{t+1}-\v{\mu}^{[0]}_{t}$ is equal to the first-order dynamics $\v{\mu}^{[1]}$ at equilibrium, this term has to be included in the computation of $\v{\dot{\mu}}^{[0]}$ to find a stationary solution during the gradient descent procedure where the gradient $\partial F/\partial\muvec$ vanishes at optimum \cite{buckley2017free}. 
Hence,
\begin{align}
\v{\dot{\mu}}^{[0]} -\v{\mu}^{[1]} =  -\frac{\partial F}{\partial \v{\mu}} = \frac{\partial\ln p(\obs_v,\v{\rho}, \v{\mu})}{\partial \v{\mu}} =  \frac{\ln p(\obs_v|\v{\mu})}{\partial \v{\mu}}+ \frac{\ln p(\v{\mu}^{[1]}|\v{\mu}^{[0]}, \v{\rho})}{\partial \v{\mu}}
\label{eq:mudot}
\end{align}

In order to compute the likelihoods, we assume that the observed image $\obs_v$ is noisy and follows a normal distribution with a mean at the value of $\v{g}(\v{\mu})$ and with variance $\v{\Sigma}_v$. Considering that every pixel contribution is independent, such as $\v{\Sigma}_v = \mathop{\mathrm{diag}}(\Sigma_{v_1},\ldots,\Sigma_{v_{h\cdot w}})$, the likelihood $p(\obs_v|\v{\mu})$ is obtained as the collection of independent Gaussians:
\begin{equation}
p(\obs_v|\v{\mu})= \prod_{k=1}^{h\cdot w}\frac{1}{\sqrt{2 \pi \Sigma_{v_{k}}}}\operatorname{exp}\left[ -\frac{1}{2 \Sigma_{v_{k}}}(x_{v_{k}}-g_{k}(\boldsymbol{\mu}))^2 \right]
\end{equation}

Analogously, the density that defines the latent variable dynamics is also assumed to be noisy and follows a normal distribution with mean at the value of the function $\v{f}(\v{\mu, \rho})$ and with variance $\v{\Sigma}_{\mu}$:
\begin{equation}
p(\v{\mu}^{[1]}|\v{\mu}^{[0]},\rho)= \prod_{i=1}^{n} \frac{1}{\sqrt{2 \pi \Sigma_{\mu_i}}} \operatorname{exp}\left[ -\frac{1}{2 \Sigma_{\mu_i}}(\mu_{i}^{[1]}-f_{i}(\v{\mu}, \v{\rho}))^2 \right]
\end{equation}

Substituting the likelihoods and computing the partial derivatives of Equation \ref{eq:mudot}, the body state is then given by the following differential equation:
\begin{align}
\label{eq:df_dmu}
    \v{\dot{\mu}}^{[0]} &= \v{\mu}^{[1]} + \underbrace{\frac{\partial \v{g}(\v{\mu})}{\partial \v{\mu}}^T}_\text{mapping}\underbrace{\v{\Sigma}_v^{-1}}_\text{precision} \underbrace{(\obs_v-\v{g}(\muvec))}_\text{prediction error}
    + \frac{\partial\v{f}(\muvec, \v{\rho})^T}{\partial\v{\mu}} \v{\Sigma}_{\mu}^{-1}(\v{\mu^{[1]}}-\v{f}({\muvec, \v{\rho}})) 
\end{align}
For notation simplicity, we name the second and the third term as $F_g$ and $F_f$ respectively.

The action is analogously computed. However, the sensory information is only a function of the action $\obs(\action)$ and therefore it only depends on the free-energy terms with sensory information $F_g$\footnote{The chain rule in quotation marks of Eq.~\ref{eq:df_da} indicates the abuse of notation. When $\obs$ is an image the gradient $\frac{\partial \obs}{\partial \v{a}}$ would be a three-dimensional tensor.}:
\begin{align}
\label{eq:df_da}
    \v{\dot{a}} \!=\! -\frac{\partial F_g}{\partial \v{a}} = -\mquote{\frac{\partial F_g}{\partial \obs}\frac{\partial \obs}{\partial \v{a}}} \!=\!
  -\frac{\partial\obs_v^T}{\partial\v{a}} \v{\Sigma}_{v}^{-1} (\obs_v\!-\!\v{g}(\v{\mu})) \!=\! -\frac{\partial\v{g}(\v{\mu})^T}{\partial \muvec} \Delta_t \v{\Sigma}_{v}^{-1} (\obs_v\!-\!\v{g}(\v{\mu}))
\end{align}
To derive the last equality, we have employed the same approximation as in \cite{oliver2019active} assuming that the actions are joint velocities. In a velocity controller scheme we can approximate the angle change between two time steps of each joint $j$ as $\partial q_j/\partial a_j= \Delta_t$, because the target values of the joint encoders $\v{q}$ are computed as $\v{q}^{t+1}=\v{q}^t + \Delta_t \action^t$ and $\Delta_t$ is a fixed value that defines the duration of each iteration. Then, assuming convergence for the sensation values at the equilibrium point with $\v{\mu} \rightarrow \v{q}$ and $\v{g}(\v{\mu}) \rightarrow \obs_v$, the term  $\frac{\partial\obs_v}{\partial\v{a}}$ can be computed using the following equation: 
\begin{equation}
\frac{\partial\obs_v}{\partial a_j}=\frac{\partial\obs_v}{\partial q_j}\frac{\partial q_j}{\partial a_j}=\frac{\partial\v{g}(\v{\mu})}{\partial \mu_j}\frac{\partial \mu_j}{\partial a_j} = \frac{\partial\v{g}(\v{\mu})}{\partial \mu_j} \Delta_t
\end{equation} 

The update rule for both $\v{\mu}$ and $a$ is finally calculated with the first-order Euler integration:
  \begin{equation}
  \begin{split}
 \v{\mu}_{t+1}=\v{\mu}_t + \Delta_t  \dot{\v{\mu}}  
 \end{split}
 \qquad \quad
  \begin{split}
        \v{a}_{t+1}=\v{a}_t + \Delta_t  \dot{\v{a}} 
        \end{split}
  \end{equation}
%

\subsection{Scaling up with deep autoencoders}
\label{sec:visual_model}
In order to perform pixel-based free-energy optimization we compute Equations~(\ref{eq:df_dmu}) and (\ref{eq:df_da}) exploiting the forward and backward pass properties of the deep neural network. We approximate the visual generative model $\v{g}(\muvec)$ and its partial derivative $\partial_{\mu} \v{g}(\muvec)$ with respect to the internal state by means of a convolutional decoder.

\subsubsection{Prediction of the expected sensation}
\begin{figure*}[hbtp!]
\centering
 \fontsize{6pt}{6pt}\selectfont
	  \hspace*{-0.6cm} 
	  \includegraphics[width=0.80\linewidth]{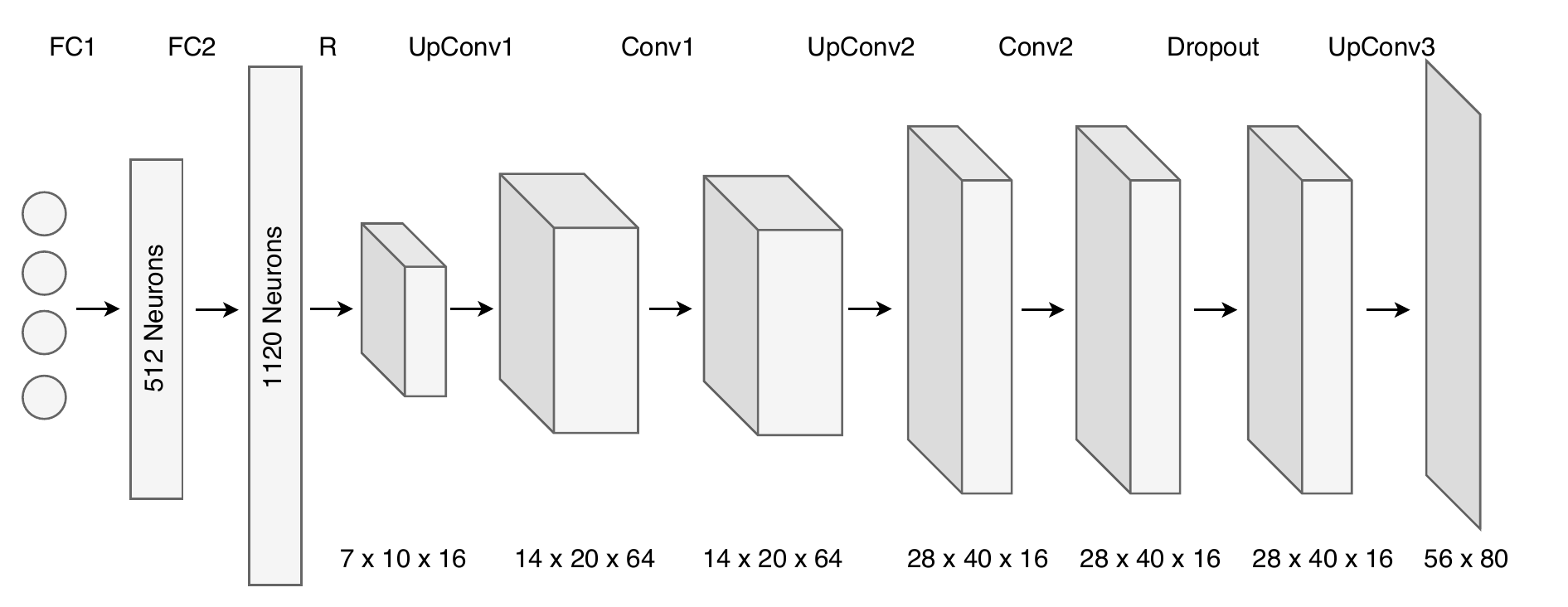}
  \caption{Network Architecture of the Convolutional Decoder (FC: Fully-connected layer, R: Reshape operator, UpConv: Transposed Convolution, Conv: Convolution)} \label{fig:conv_decoder}
\end{figure*}

We approximate the forward model $\v{g}(\muvec)$ by means of a generative network, based on the architecture proposed in~\cite{dosovitskiy2016learning}, described in Fig.~\ref{fig:conv_decoder}. It outputs the predicted image given the robot's n-dimensional internal belief of the body state $\v{\mu}$, e.g., the joint angles of the robot.

The input goes through 2 fully-connected layers (FC1 and FC2). Afterwards, the transposed convolution (UpConv) is performed to upsample the image. This deconvolution uses the input as the weights for the filters and can be regarded as a backward pass of the standard convolution operator~\cite{dumoulin2016guide}. Following \cite{dosovitskiy2016learning}, each transposed convolution layer was followed by a standard convolutional layer, which helps to smooth out the potential artifacts from the transposed convolution step. There is an additional 1D-Dropout layer before the last transposed convolution layer to avoid overfitting and achieve better generalization performance. All layers use the rectified linear unit (ReLU) as the activation function, except for the last layer, where a sigmoid function was used to get pixel intensity values in the range $[0,1]$. Throughout the consecutive UpConv-Conv operations in the network, the number of channels is increased and decreased again to get the required output image size.

\subsubsection{\textbf{Backward pass and mapping to the latent variable}}
\label{sec:bakward}

An essential term, for computing both perception and action, is the mapping between the error in the sensory space and the inferred variables: $\frac{\partial \v{g}(\v{\mu})}{\partial \v{\mu}}$. This term is calculated by performing a backward pass over the convolutional decoder. In fact, we can compute the whole partial derivative of the visual input term $\partial F_{g}/\partial \muvec$ just in one forward and backward pass. The reason is that the prediction error between the expected and observed sensation $(\obs_v-\v{g}(\muvec))$ multiplied by the inverse variance and the partial derivative is equivalent to applying the backpropagation algorithm. It is important to note that when the function $\v{g}(\muvec)$ outputs images of size $w \times h$, $\frac{\partial \v{g}(\v{\mu})}{\partial \v{\mu}}$ is a three dimensional tensor. We stack the output into a vector $ \in \mathbb{R}^{w\cdot h}$ (row major). The following equation is obtained:

\begin{equation} \label{eq:derv_exp}
\frac{\partial F_{g}}{\partial \v{\mu}}=
\underbrace{\begin{bmatrix}
\frac{\partial g_{1,1}}{\partial \mu_1}       &   \frac{\partial g_{1,2}}{\partial \mu_1}       & \dots     &   \frac{\partial g_{w, h}}{\partial \mu_1}       \\
\frac{\partial g_{1,1}}{\partial \mu_2}       &   \frac{\partial g_{1,2}}{\partial \mu_2}       & \dots     &   \frac{\partial g_{w, h}}{\partial \mu_2}       \\
\vdots  &  \vdots   &   \vdots  &   \vdots  \\   
\frac{\partial g_{1,1}}{\partial \mu_4}       &   \frac{\partial g_{1,2}}{\partial \mu_4}       & \dots     &   \frac{\partial g_{w, h}}{\partial \mu_4}       \\
            \end{bmatrix}
            }_{\left(\frac{\partial \v{g}}{\partial \v{\mu}}\right)^T}
\underbrace{\begin{bmatrix}
\frac{\partial F_g}{\partial g_{1,1}}     \\
\frac{\partial F_g}{\partial g_{1,2}}      \\
\vdots  \\
\frac{\partial F_g}{\partial g_{w, h}}     \\
            \end{bmatrix}
            }_{\frac{\partial F_g}{\partial \v{g}}}
\end{equation}
where $-\frac{\partial F_{g}}{\partial g_{i,l}}$ is given by $\frac{1}{\Sigma_{v_{i,l}}}(x_{v_{i,l}}-g_{i,l}(\muvec))$. The action is computed by reusing this term and multiplying it by $\Delta_t$.

\subsection{\textbf{Formalizing the task with attractor dynamics}}
\label{sec:attractor_dyn}
In \textit{Active Inference} we include the goal as prior in the body state dynamics function $\v{f}(\muvec,\v{\rho})$. For example, to perform a reaching task, we encode the desired goal of the robot as an instance in the sensory space (image), which acts as an attractor generated by the causal variable $\v{\rho}$. This produces an error in the inferred state that will promote an action towards the goal. Note that the error $\v{\rho}-\v{g}(\muvec)$ is zero when the prediction matches the desired goal. We define the body state dynamics with a causal variable attractor as: 
\begin{equation}
\label{eq:attractor}
\v{f}(\v{\mu}, \v{\rho}) = \v{T}(\v{\mu}) \beta \left(\v{\rho}-\v{g}(\v{\mu}) \right),
\end{equation}
where $\beta$ is a gain parameter that defines the intensity of the attractor and $\v{T}(\muvec)= \partial \v{g}(\muvec)^T/\partial \muvec$ is the mapping from the sensory space (e.g. pixel-domain) to the internal belief $\muvec$ (e.g. joint space). Note that this term is obtained through the backward pass of the decoder. Finally, substituting in Eq.~(\ref{eq:df_dmu}) the new dynamics generative model we write the last term $\partial F_f/\partial \muvec$ as:
\begin{align}
    -\frac{\partial F_f}{\partial \muvec} =
    \frac{\partial\v{f}(\muvec, \v{\rho})^T}{\partial\v{\mu}} \v{\Sigma}_{\mu}^{-1}\left( \v{\mu^{[1]}}- \frac{\partial \v{g}(\muvec)^T}{ \partial \muvec} \beta \left(\v{\rho}-\v{g}(\v{\mu}) \right)\right)
\end{align}

In the final model used in the experiments, we have further simplified this equation by not including the first-order internal dynamics into the optimization process $\muvec^{[1]}=\v{0}$ and noting that the correct mapping and direction from the sensory space to the latent variable is already provided by $\partial \v{g}(\muvec)^T / \partial \muvec$. Thus, we greedily approximate $\partial \v{f}(\muvec,\v{\rho})/\partial\muvec$ to $-1$,  avoiding the Hessian computation of $\v{T}$ but introducing an optimization detriment. With these assumptions, the partial derivative of the dynamics term becomes:
\begin{align}
    -\frac{\partial F_f}{\partial \muvec} = 
     \v{\Sigma}_{\mu}^{-1}\left( \frac{\partial \v{g}(\muvec)^T}{ \partial \muvec} \beta \left(\v{\rho}-\v{g}(\v{\mu}) \right)\right) = \v{\Sigma}_{\mu}^{-1} \v{f}(\muvec,\v{\rho}) 
\end{align}

\subsection{PixelAI algorithm}

Algorithm \ref{alg:deep_active_alg} summarizes the proposed method. In the robot body perception and action application, $\obs_v$ is set to the image provided by the robot monocular camera and the decoder input becomes the robot proprioception (e.g. joint angles). The convolutional decoder is trained with this mapping obtaining a predictor of the visual forward model. The prediction error $\v{e_v}$ is the difference between the expected visual sensation and the observation (line 6). The variational free-energy optimization, for perception (line 7) and action (line 8), updates the differential equations that drive the state estimation and control. Finally, we added in the dynamics term the possibility of inputting desired goals in the visual space (line 13). Although this implementation assumes that $\muvec^{[1]}=\v{0}$, it is straight forward to add the 1st order dynamics when there is velocity image information or joint encoders~\cite{oliver2019active}.

\small
\begin{algorithm}[hbtp!]
\caption{PixelAI: Deep Active Inference Algorithm}
\label{alg:deep_active_alg}
\begin{algorithmic}[1]
\Require $\v{\Sigma}_v, \v{\Sigma}_\mu, \beta, \Delta_t$
\State $\mu  \gets \text{Initial joints angle estimation}$
\While{(true)}
    
    \State $\obs_v \gets Resize(\text{camera image})$\Comment{Visual Sensation}
    
    \State $\v{g}(\v{\mu}) \gets  ConvDecoder.\text{forward}(\v{\mu})$ 
    \State $\partial \v{g} \gets  ConvDecoder.\text{backward}(\v{\mu})$
    \State $\v{e}_v = (\obs_v  - \v{g}(\v{\mu}))$ \Comment{Prediction error}
    \State $\dot{\muvec} = K_{\v{\Sigma}_v}\partial \v{g}^T \v{e}_v/\v{\Sigma}_v$ \footnotemark
    \State $\dot{\action} = - (\partial \v{g}^T \v{e}_v/\v{\Sigma}_v)\Delta_t$
    \If{$\exists \v{\rho} $} \Comment{Goal attractor $\rho$ dynamics}
        \State $\v{e}_f = \beta(\v{\rho}-\v{g}(\muvec))$  
        \State $\dot{\muvec} = \dot{\muvec} + \partial \v{g}^T \v{e}_f/\v{\Sigma}_\mu$
    \EndIf
    \State $\muvec = \muvec + \Delta_t \dot{\muvec}$;  $\action = \action + \Delta_t \dot{\action}$ \Comment{1st order Euler integration}

    \State SetVelocityController($a$) 
\EndWhile
\end{algorithmic}
\vspace{-2px}
\end{algorithm}
\normalsize
\footnotetext{The gain parameter $K_{\v{\Sigma}_v}$ is added to allow the model to generate large action values without increasing the internal belief increments.}

\section{Experiments}
\label{sec:experimentalsetup}

\begin{figure}[hbtp!]
	\centering
	\subfigure[Simulated Nao]{
		\centering
		\includegraphics[width=0.3\linewidth]{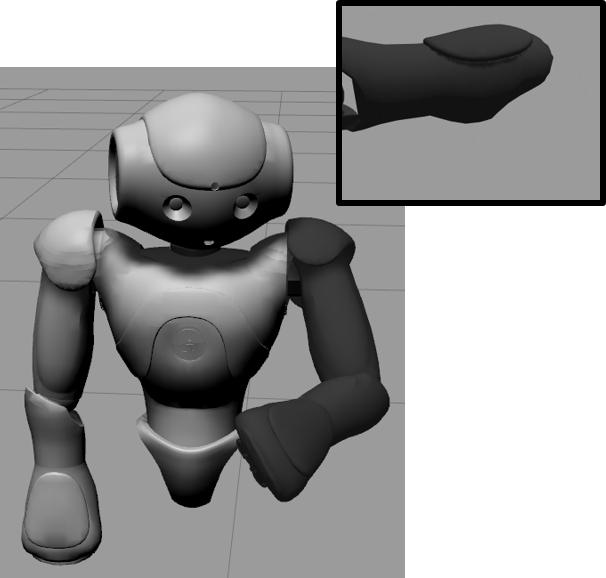}
		\label{fig:gazebo_nao}
	}
	\subfigure[Real Nao]{
		\centering
		\includegraphics[width=0.3\linewidth]{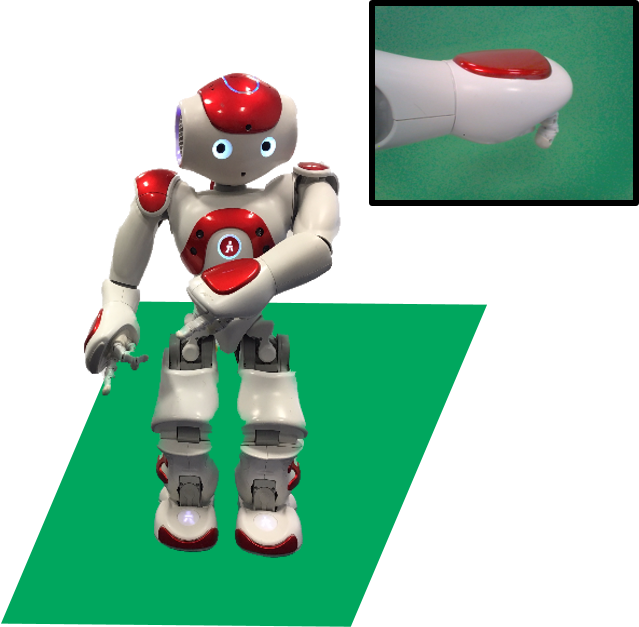}
		\label{fig:real_nao}
	}
	\label{fig:setup}
	\caption{Experimental setup in simulation (Gazebo) and with the real robot.}
\end{figure}

We tested the PixelAI in both simulated and real Aldebaran NAO humanoid robot (Fig.~\ref{fig:setup}). We used the left arm to test both perception and action schemes. The dataset and the code to replicate the experiments can be found in \url{tobereleased}.



\subsection{Visual forward model}
\paragraph{\textbf{Data acquisition \& preprocessing}.}
\label{sec:conv_decoder_data}
The dataset used to train the model consisted of 3200 data samples of the left arm elbow and shoulder joint readings ($\v{q}= [q_1, q_2, q_3, q_4]^T$) and the observed images $\obs_v$ obtained through NAO's bottom camera\footnote{In simulation, the color of the right arm was changed to dark grey to achieve contrast with the grey background in the camera images.}. Data samples were generated using three different methods and later concatenated ($\sim25\%$, $\sim20\%$ and $\sim55\%$ for each method).


In the first method, the joint angles were randomly drawn from a uniform distribution in the range of the joint limits. Afterwards, it required the manual elimination of samples where the robot's arm was out of the camera frame. The ratio of the acquired images with the robot hand centered in the camera image was significantly lower than the images with the hand located at the corners of the frame. To reduce this drawback, in the second method, the robot's arm was manually moved by an operator and the joint angle readings were recorded during these trajectories. This way, a subset of data was obtained, where the robot hand was centered in the image. Finally, in the third method, a multivariate Gaussian was fit to the second subset using the expectation-maximization algorithm and random samples were drawn from this Gaussian for the third and final part of the dataset. The goal was to introduce randomness to the centered-images and not be limited to the operator's choice of trajectories.

For the images collected in the Gazebo NAO Simulator, the only preprocessing step performed was re-sizing the image of size $640 \times 480$ to $80 \times 56$. For the real NAO, the images were obtained on a green background (Fig.~\ref{fig:real_nao}) and the following preprocessing steps were performed: 1) median filtering with kernel size 11 on the original image, 2) masking the monochrome background,e.g. green, in the HSV color space and replacing it with dark gray to ensure contrast 3) converting the image to grayscale, and 4) resizing image to dimensions $80 \times 56$.

\paragraph{\textbf{Training}.}
The convolutional decoder was trained using the ADAM optimizer using a mini-batch of size 200 samples and an initial learning rate of $\alpha=10^{-4}$ with exponential decay of $0.95$ every 5000 steps. The training was stopped after ca. 7000 iterations for the simulated NAO dataset and 12000 iterations for the real NAO dataset to avoid overfitting, as the test set error started to increase for the corresponding model. The output of the second fully connected layer (FC2) was an 1120-dimensional vector that was reshaped into a $7 \times 10 \times 16$ tensor. The UpConv layers all use stride equal to 2 and a padding of 1. Moreover, a kernel with a size of $4 \times 4$ was chosen to avoid checkerboard artifacts due to uneven overlap~\cite{odena2016deconvolution}. Convolutional layers used a kernel size 3, stride 1 and padding 1. The ratio of drop-out was set to 0.15. The final layer outputs a $1 \times 56 \times 80$ image corresponding to a grayscale image.

\subsection{Benchmark for PixelAI}
A benchmark with three levels of difficulty was created to evaluate the performance of PixelAI on randomized samples for both perceptual and active inference. A set of 50 different cores (i.e. images of the arm) were generated by sampling the multivariate Gaussian distribution (see method 2 in section \ref{sec:conv_decoder_data}). A subset of the generated cores is shown in Fig.~\ref{fig:cores}. For each of the cores, 10 different random tests were performed. In total, there were 2500 trials composed of 5 runs of 500 testing image arm poses per benchmark level \footnote{For the real robot benchmark tests (perceptual inference) only a subset of 20 cores were used.}. The test samples for each core were generated differently depending on the benchmark level:
\begin{itemize}
\item Level 1 (close similar poses): One of the 4 joints was chosen randomly and a random perturbation $\pm[5^{\circ}, 10^{\circ}]$ sampled from a uniform distribution was added to the joint angle value to generate the new test sample. 
\item Level 2 (far similar poses): For all of the 4 joints, a random perturbation $\pm[5^{\circ}, 10^{\circ}]$ was sampled from a uniform distribution and added to the core joint angles. 
\item Level 3 (random): For each core, 10 different cores were chosen randomly and used as the test samples.
\end{itemize}

\paragraph{\textbf{Perceptual Inference}.}
In order to evaluate the body perception performance, the robot has to infer its real arm pose just using visual information. The robot's arm was initialized to each core pose and then 10 separate test runs were performed, where the internal belief was set to a perturbed value of the corresponding pose. These tests are static in nature, i.e. the change solely takes place in the internal predictions of the robot. The goal is that the robot internal belief $\v{\mu}$ converges to the true arm position, which is equal to the joint angles of the chosen core. 

\paragraph{\textbf{Active Inference}.}
In order to evaluate the perception and action performance, the core poses were treated as the desired goal (image) encoded as an attractor in the model. Again for each core, 10 separate test runs were performed. In this case, the robot arm was initialized to a core pose and also the initial internal belief was set to the current joint measurements: $\v{\mu}=\v{q}$. In each test, the goal was that the robot reached the desired imagined arm pose. The update of the internal belief should generate an action to compensate for the mismatch between the current and the predicted visual sensations. In a successful test run, the robot arm should move to the imagined arm position and the internal belief should also converge to the imagined joint angles, so that: $\obs_v = \v{g}(\v{\mu}) = \v{\rho}$. 

\paragraph{{Algorithm Parameters}.}
The parameters of the Pixel AI algorithm are $\v{\Sigma}_v, K_{\v{\Sigma}_v}, \beta, \Delta_t$, which were determined empirically \footnote{Since $\beta$ and $\v{\Sigma}_{\mu}$ achieve the same effect, $\v{\Sigma}_{\mu}$ is always set to 1 and only $\beta$ is kept as a parameter, eliminating redundancy.}. The intuition behind the variance terms is as follows: the prediction errors get multiplied by the inverse of the variances so these actually weigh the relevance of the corresponding sensory information error~\cite{buckley2017free}. The $\beta$ term, that is part of the attractor dynamics (see Eq. \ref{eq:attractor}), essentially has the same effect and it controls how much we want to push the internal belief in the direction of the attractor. Finally, the $\Delta_t$ depends on the internal time of the robot loop execution. The parameters used for benchmark levels 1 and 2 were the same. For level 3 in perceptual inference, we used a smaller $\v{\Sigma}_v$ until the visual prediction error was below a certain threshold ($0.01$). This introduces a new model parameter $\gamma_{\v{\Sigma}_v}$ which is used to scale $\v{\Sigma}_v$, once the error threshold is reached. This heuristic method of adaptation helped speed up the convergence for the more complex level 3 trajectories. Finally, the generated actions (velocity values) were clipped so that each joint could not move more than $[-2^\circ; 2^\circ]$ each time step. 


\begin{table}[htpb!]
  \caption{PixelAI Parameters used for the perceptual and active inference benchmarks in simulation.}
  \label{param_table}
  \centering
  \begin{tabular}{@{}lcccc@{}}
    \toprule
         & $\v{\Sigma}_v$ & $\beta$  & $K_{\v{\Sigma}_v}$ & $\gamma_{\v{\Sigma}_v}$ \\ 
    \midrule
     \textbf{Active Inference}\\
        Level 1  & $6\times 10^{2}$ & $2\times 10^{-5}$    & $10^{-3}$  & 1 \\
        Level 2  & $2\times 10^{2}$    & $5\times 10^{-5}$    & $10^{-3}$  & 1 \\
        Level 3  &  $20$   &  $5\times 10^{-4}$   & $10^{-3}$  & 1 \\
      \midrule
         \textbf{Perceptual Inference}\\
        Level 1  & $2\times 10^{4}$ &  -  &  -   &  1\\
        Level 2  & $2\times 10^{4}$ &  -  &  -   & 1 \\
        Level 3  & $2\times 10^{3}$ &  -  &  -   &  10\\
      \bottomrule         
  \end{tabular}
\end{table}

\section{Results}
\label{sec:results}
\subsection{Statistical analysis of perception and action in simulation}
\begin{figure}[hbtp!]
\includegraphics[width=0.95\linewidth]{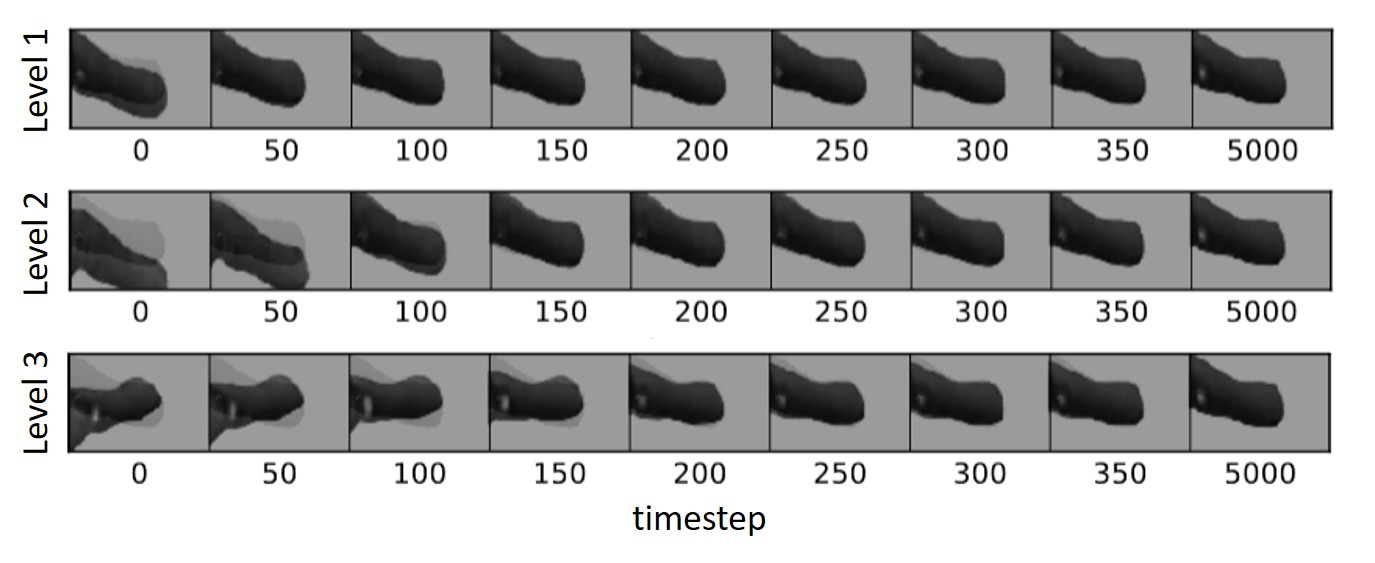}
	\label{fig:internal_traj}
	\vspace{-0.5cm}
	\caption{Example of the internal trajectories of the latent space during the perceptual inference tests for three different levels of difficulty for core 3.}
\end{figure}

\begin{figure*}[t!]
	\centering
	\subfigure[Benchmark cores subset]{
		\centering
		\includegraphics[width=0.27\textwidth]{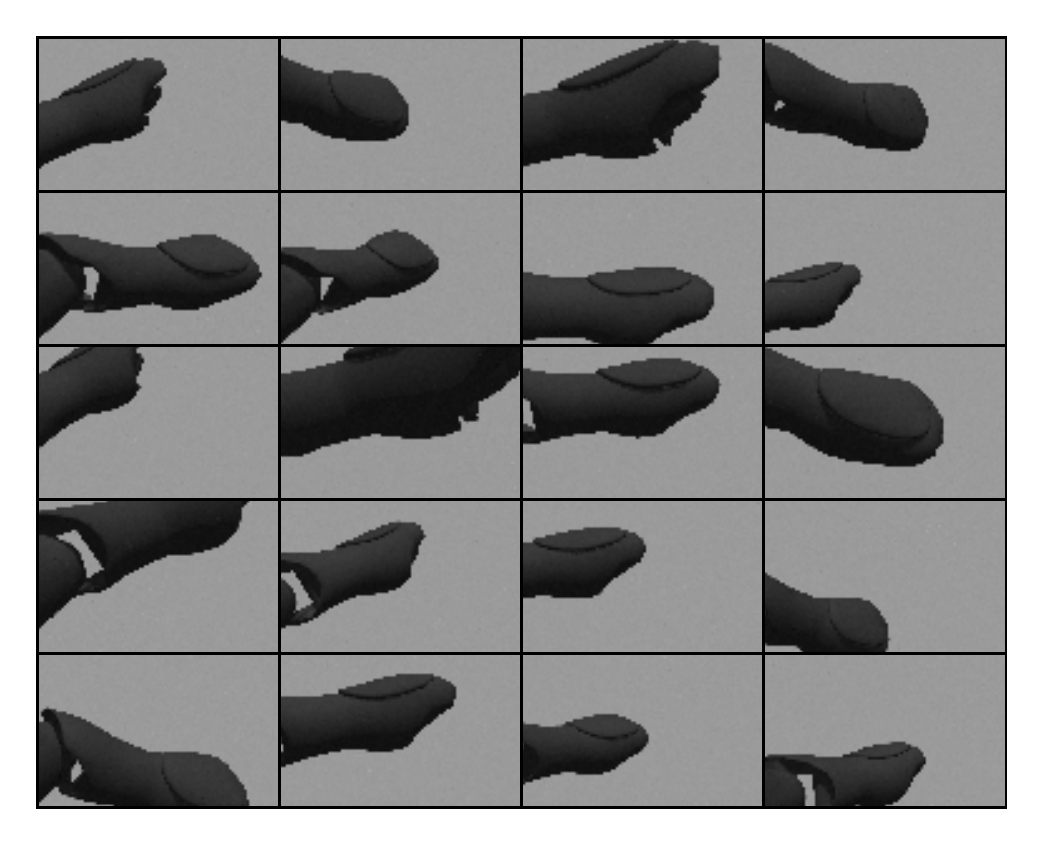}
		\label{fig:cores}
	}
	\subfigure[Visual Prediction Error]{
		\centering
		\includegraphics[width=0.3\textwidth]{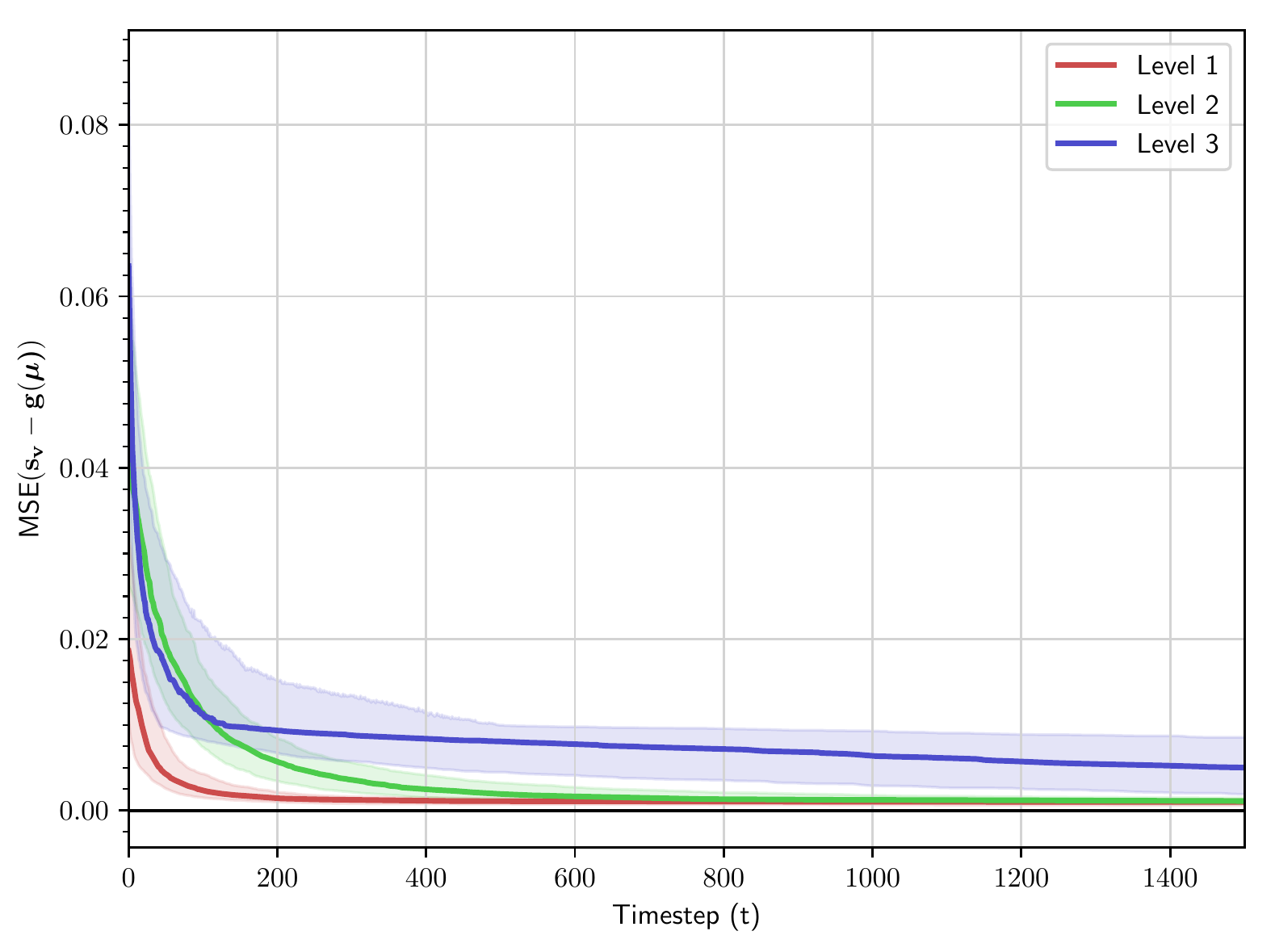}
		\label{fig:sim_perception_vis}
	}
	\subfigure[L2 Norm of $\v{s_p}-\v{\mu}$.]{
		\centering
		\includegraphics[width=0.3\textwidth]{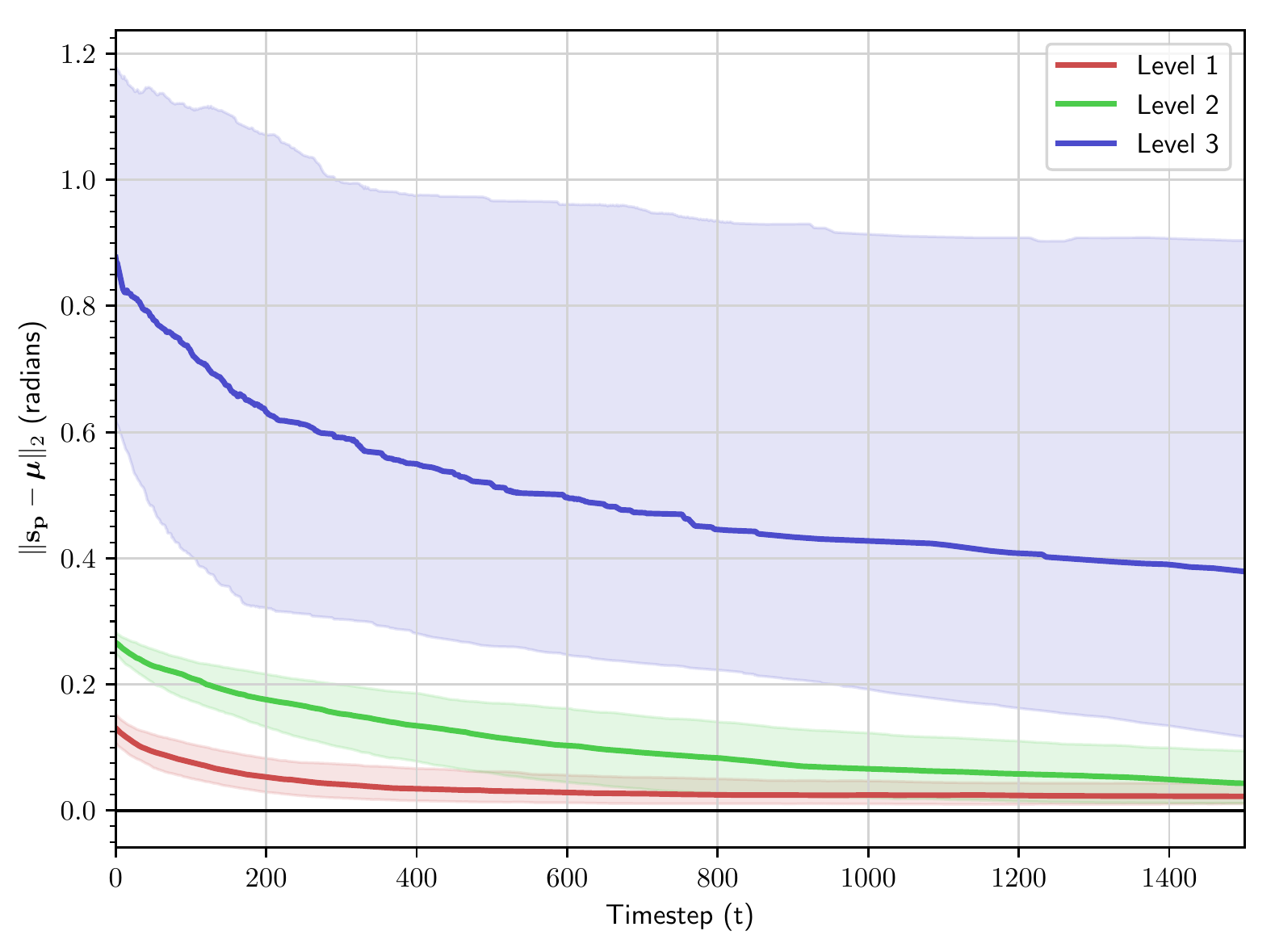}
		\label{fig:sim_perception_mu}
	}	
	\caption{(a): A subset of the cores used for the benchmarking tests. (b)-(c): Simulated Nao perceptual inference results for all levels (1-3) of the benchmark are shown. The L2-Norm of the error between internal belief $\v{\mu}$ and $\mathbf{s_{p}}$ is plotted, as well as the visual prediction error.}
	\label{fig:sim_perception}
\end{figure*}

First, perceptual inference tests were run for 5000-time steps for all 3 levels. An example of the perceptual inference for each level is depicted in Fig.~\ref{fig:internal_traj}. For level 1 and 2, with less than 150 iterations, the algorithm converged to the ground truth, while inferring the body location from a totally random initialization (level 3) raised considerably the complexity. Table \ref{table:mae_sim2} shows the resulting average of all trials of the mean absolute joint errors ($|\v{q}_{true}-\muvec|$). Level 1 and 2 results converged to internal belief values successfully. Figure~\ref{fig:sim_perception_mu} shows the error during the optimization process. Shoulder pitch and shoulder roll angles were estimated with better accuracy compared to the elbow angles. This is due to the fact that a small change in the shoulder pitch angle yields to a greater difference in the visual field in comparison with the same amount of change in the elbow roll angle. Since PixelAI achieves perception by minimizing the visual prediction error, the accuracy increases when the pixel-based difference is stronger. Therefore, the mean error and standard deviation increase for the elbow joint angle estimations.

\begin{figure*}[t!]
	\centering
	\subfigure[True arm position (camera image).]{
		\centering
		\includegraphics[width=0.23\textwidth]{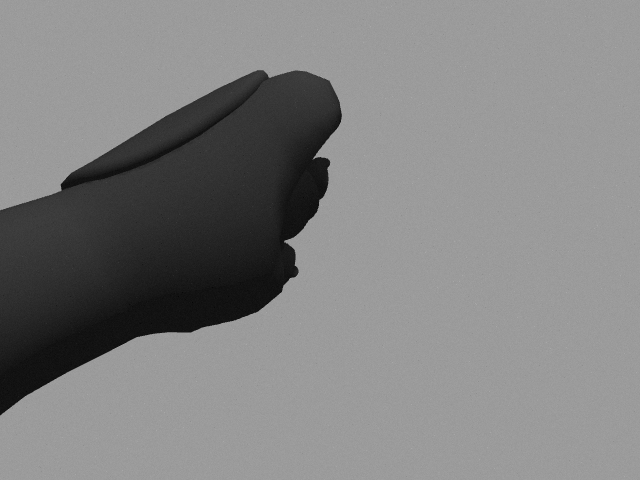}
		\label{fig:true_local_min}
	}
	\subfigure[Initial internal belief deep convolutional decoder output $\v{g}(\v{\mu})$ (overlaid on the camera image).]{
		\centering
		\includegraphics[width=0.25\textwidth]{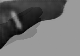}
		\label{fig:beginning_local_min}
	}
	\subfigure[Deep convolutional decoder output $\v{g}(\v{\mu})$ for end internal belief (overlaid on the camera image).]{
		\centering
		\includegraphics[width=0.25\textwidth]{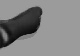}
		\label{fig:end_local_min}
	}	
    \\
    \subfigure[Visual prediction error.]{
		\centering
		\includegraphics[width=0.3\textwidth]{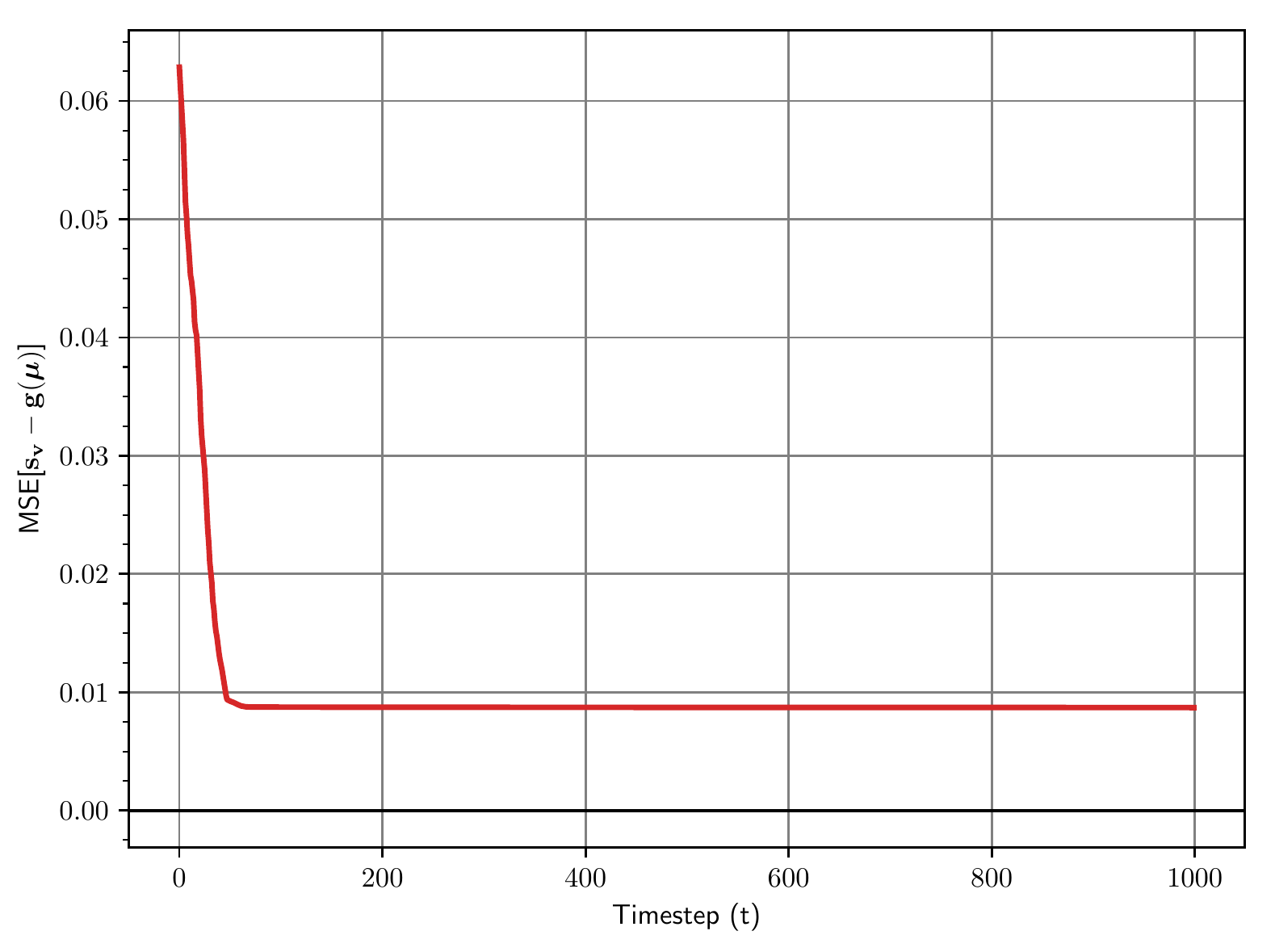}
		\label{fig:viserr_level_lm12}
	}
	\subfigure[L2 Norm of $\v{\mu}-\v{s_p}$.]{
		\centering
		\includegraphics[width=0.3\textwidth]{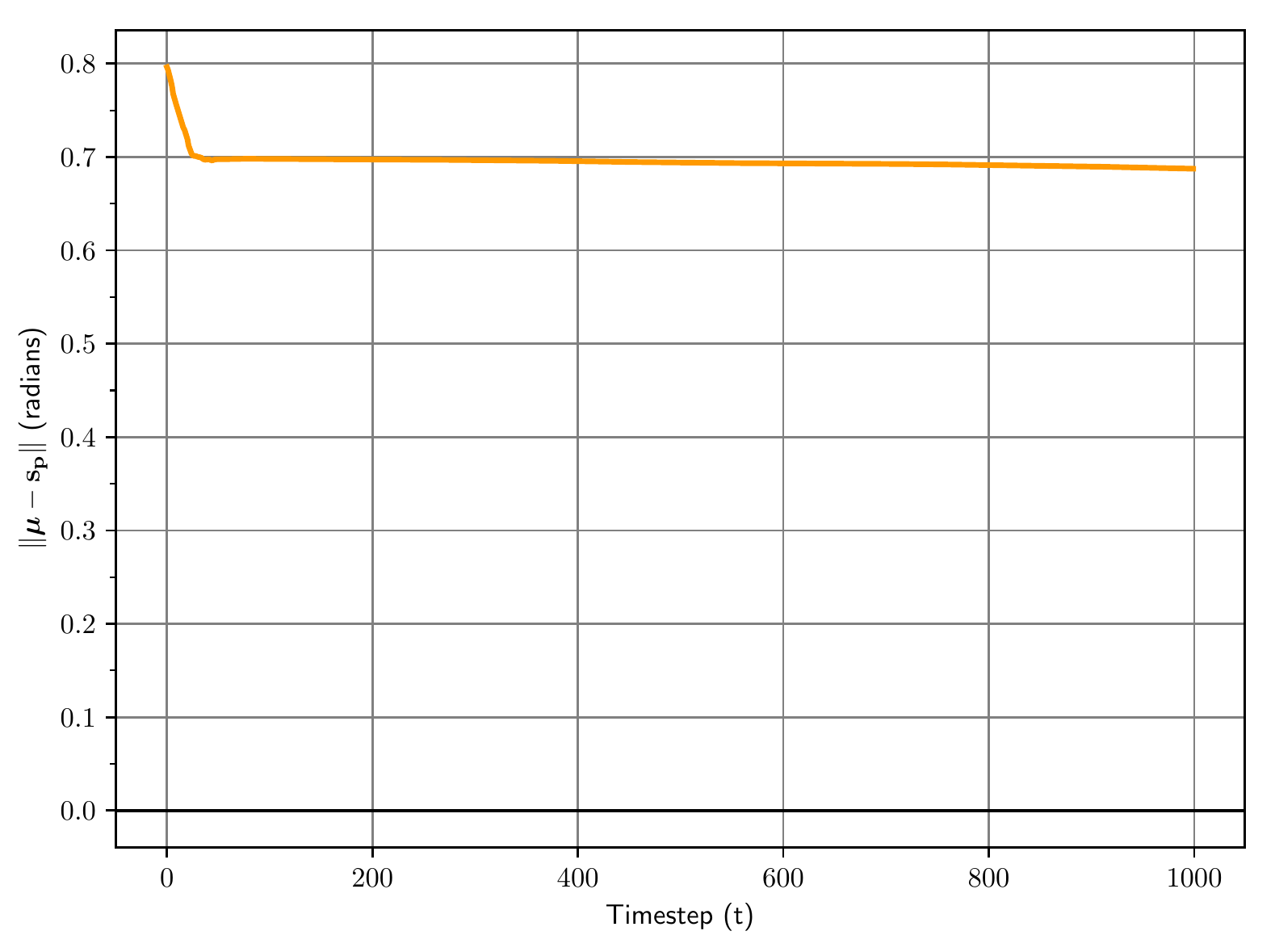}
		\label{fig:mu_err_local_min}
	}	
	\caption{Example of the local minima problem in level 3 perceptual inference tests.}
	\label{fig:local_minima}
\end{figure*}

The errors in level 3, where the robot had to converge to random arm locations, were larger compared to levels 1 and 2, as shown in Fig.~\ref{fig:sim_perception_mu}. This is due to two reasons. The first one is the local minima problem inherent of our gradient descent approach. The second one affects the desired joint position, as depicted in Fig.~\ref{fig:local_minima}: several joint solutions have small visual prediction error increasing the risk of getting into a local minimum.

\begin{table}[htpb!]
  \caption{Perceptual inference joint angles mean absolute error (in degrees).}
  \label{table:mae_sim2}
  \centering
  \begin{tabular}{lllll}
    \toprule
    Level  & Shoulder Pitch   & Shoulder Roll    & Elbow Yaw        & Elbow Roll        \\  
    \midrule 
    1      & 0.259 $\pm$0.343 & 0.333 $\pm$0.408 & 0.848 $\pm$0.956 & 0.940 $\pm$1.061 \\
    2      & 0.387 $\pm$0.711 & 0.617 $\pm$0.978 & 1.185 $\pm$1.475 & 1.567 $\pm$2.184  \\
    3      & 5.316 $\pm$11.057 & 5.609 $\pm$7.679 & 17.64 $\pm$26.08 & 12.25 $\pm$15.41  \\
    \bottomrule
  \end{tabular}
\end{table}

\begin{figure*}[t!]
	\centering
	\subfigure[Visual Prediction Error]{
		\centering
		\includegraphics[width=0.3\textwidth]{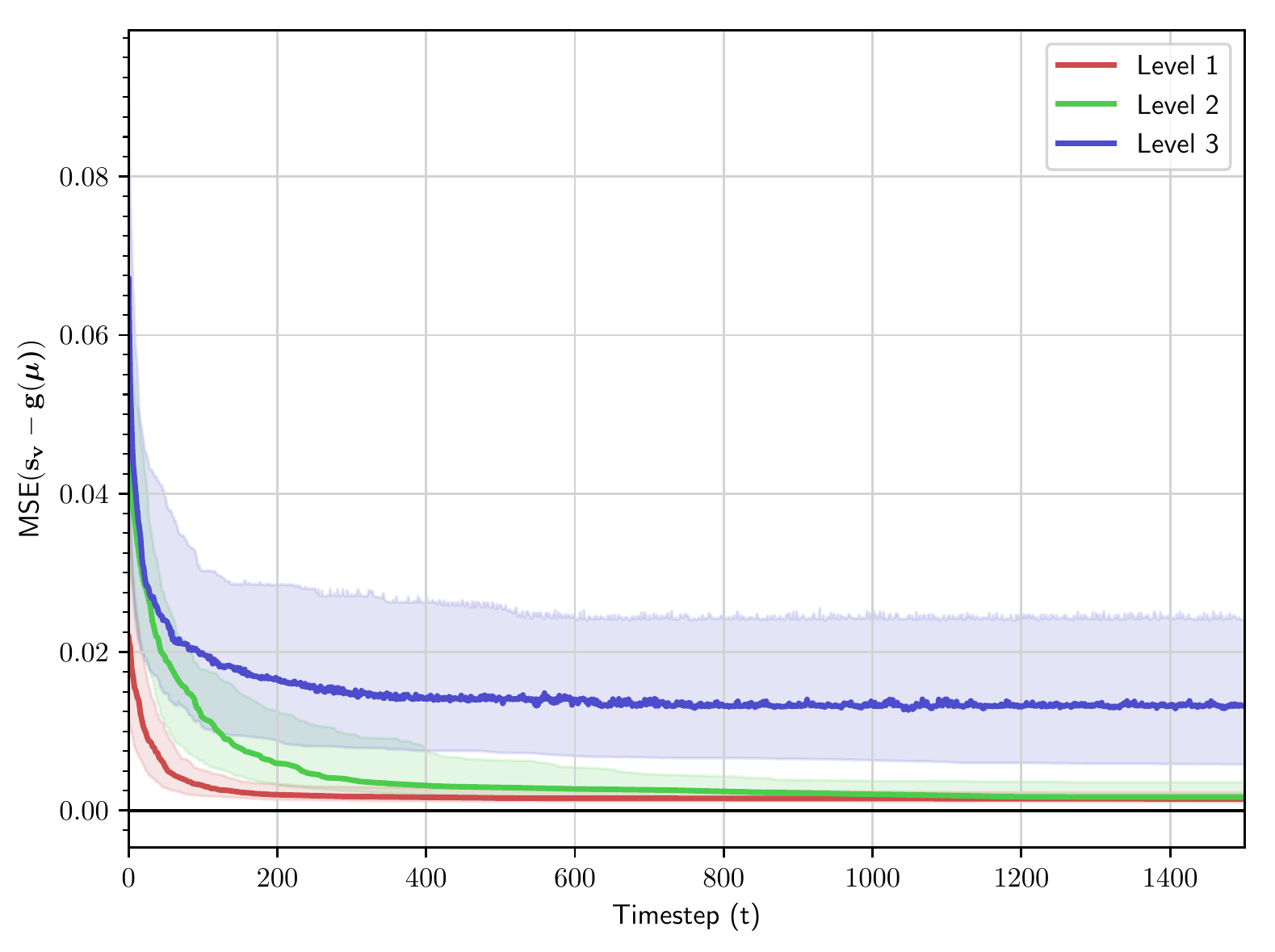}
		\label{fig:vis_err_all_real}
	}
	\subfigure[L2 Norm of $\v{\mu}-\v{s_p}$]{
		\centering
		\includegraphics[width=0.3\textwidth]{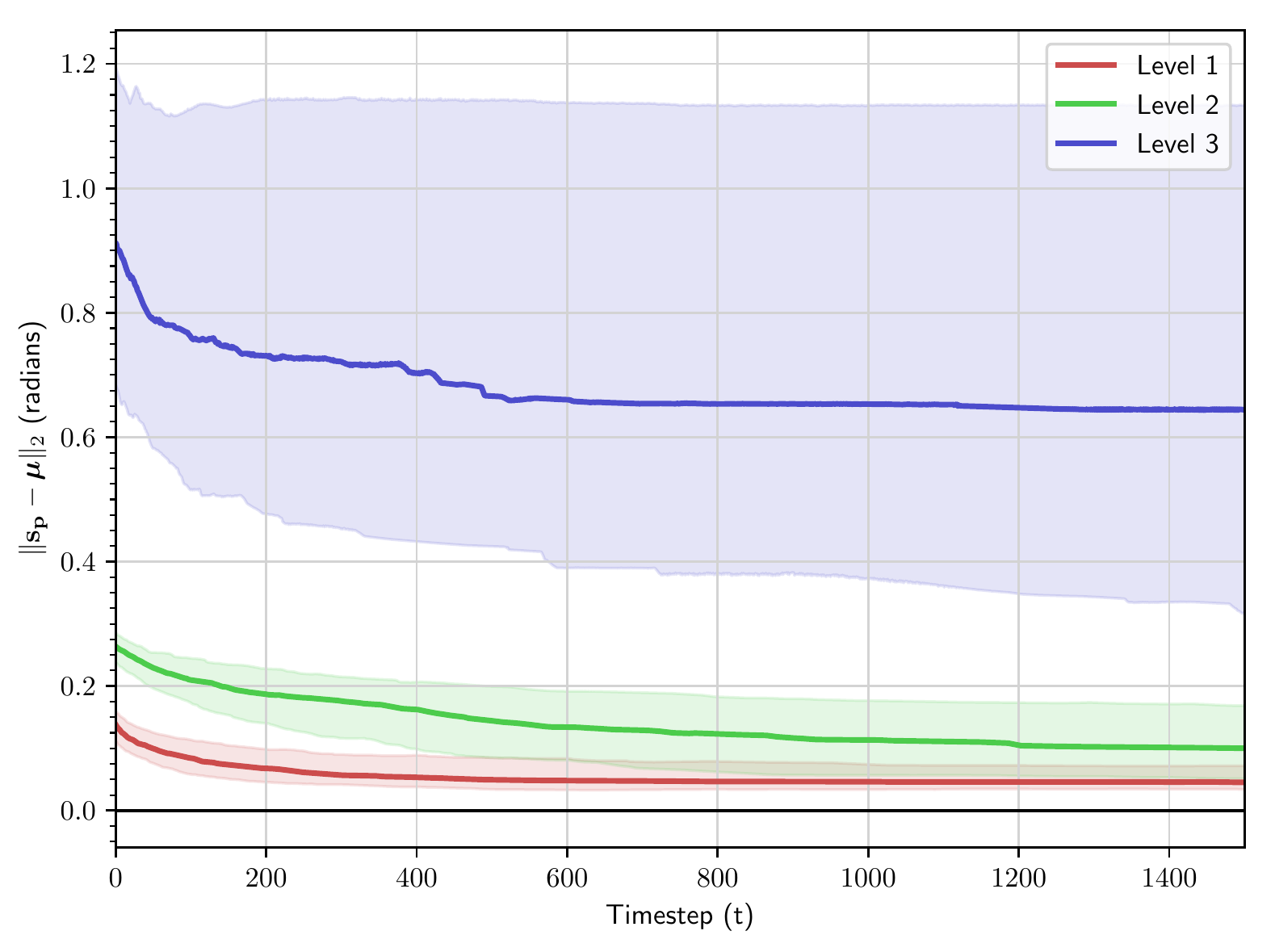}
		\label{fig:mu_err_all_real}
	}	
	\caption{Real NAO perceptual inference results for all 3 benchmark levels}
	\label{fig:all_levels_real}
\end{figure*}

\begin{figure*}[t!]
	\centering
	\subfigure[Visual error]{
		\centering
		\includegraphics[width=0.45\textwidth]{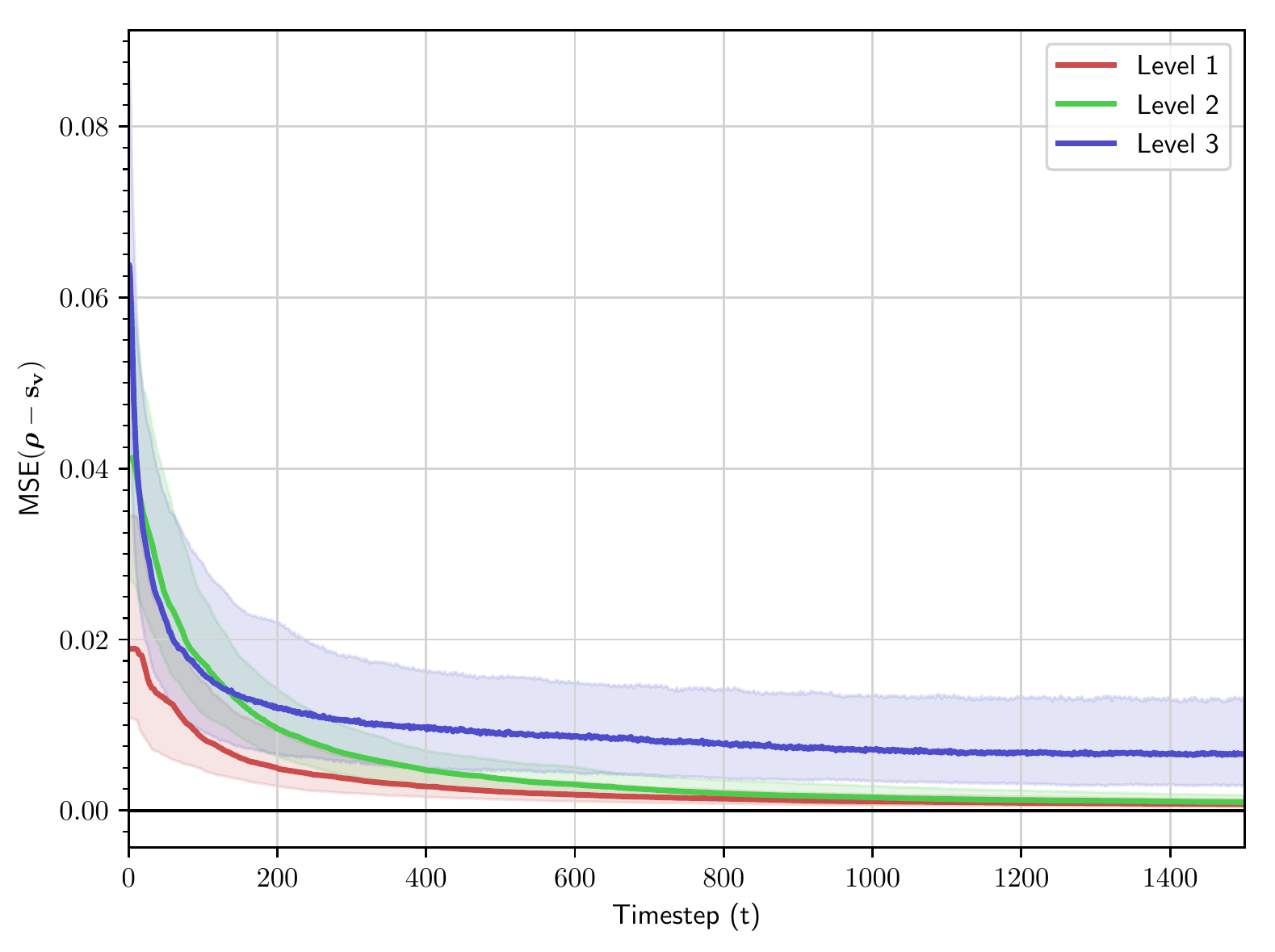}
		\label{fig:active_vis_err}
	}
	 \subfigure[L2-Norm of Joint Angle Errors]{
		\centering
		\includegraphics[width=0.45\textwidth]{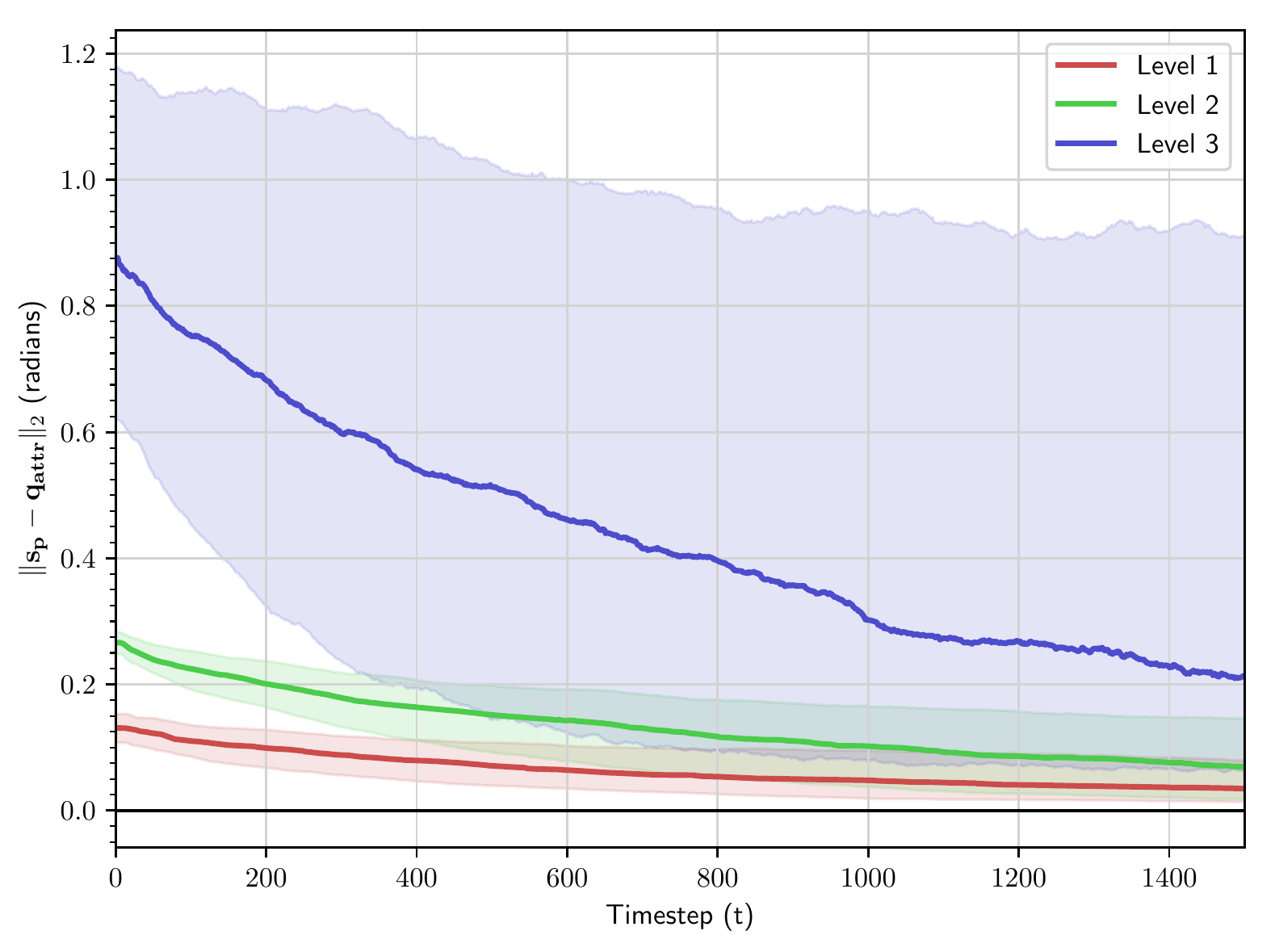}
		\label{fig:active_q_err}
	}
	\caption{Simulated NAO active inference test results for all three levels: (a) Visual error between the visual attractor $\v{\rho}$ and the observed camera image $\v{s_v}$. (b) L2-Norm of the error between the joint angles of the attractor position $\v{q_{attr}}$ and the proprioceptive sensor readings $\v{s_p}$}.
	\label{fig:active_level1_2}
\end{figure*}

Secondly, active inference tests with goal images were performed using the simulated NAO for 1000 time steps in the benchmark levels. The results for levels 1 and 2 are shown in Fig.~\ref{fig:active_level1_2}. The joint encoder readings followed the internal belief values through the actions generated by free-energy optimization. Level 3 performance detriment shows that interacting is more complex than perceiving as it includes the body and the world constraints.


\begin{figure*}[t!]
	\centering
	\subfigure[Level 2: Visual error]{
		\centering
		\includegraphics[width=0.45\textwidth]{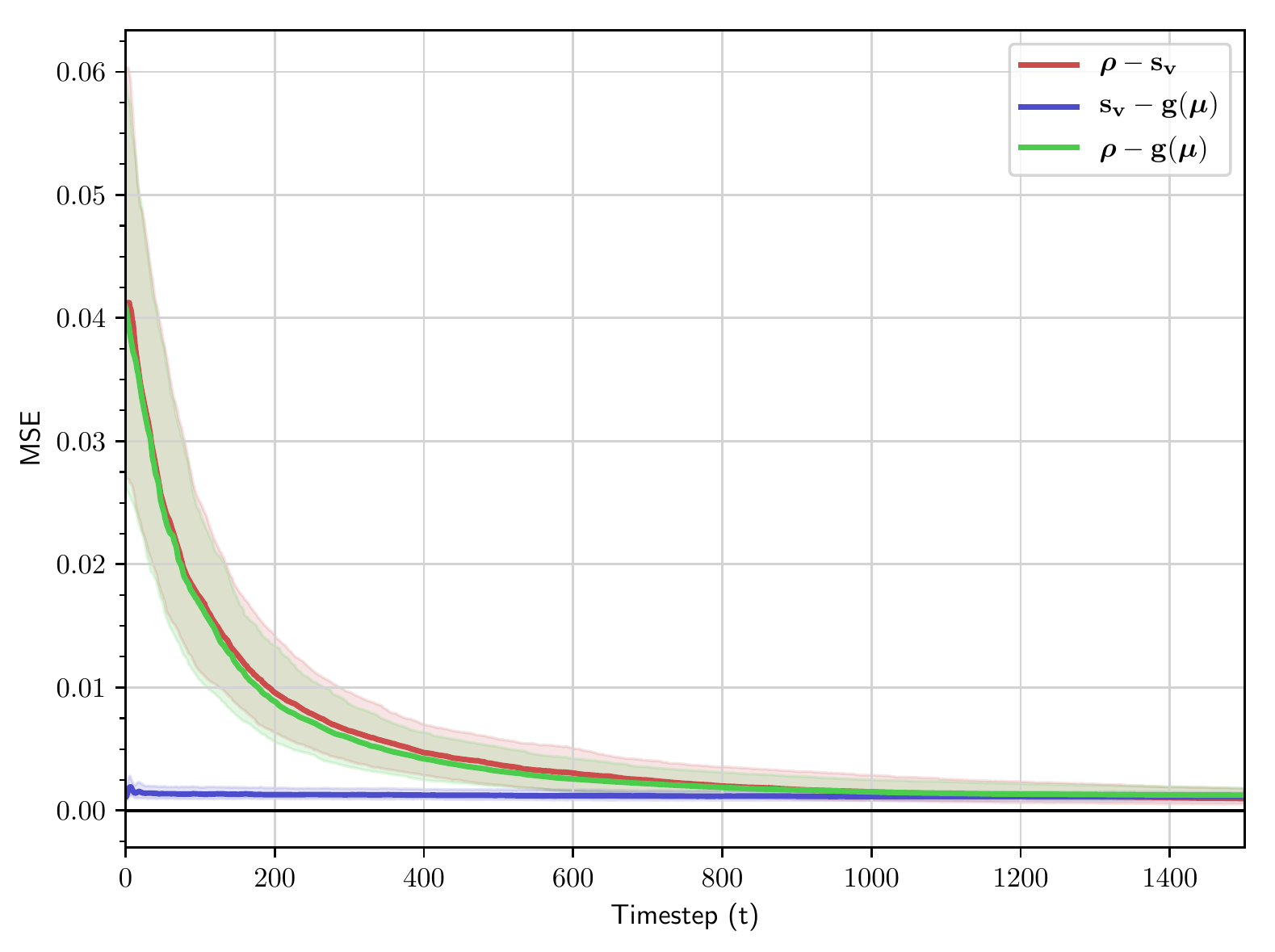}
		\label{fig:viserr_active_level3}
	}	
	\subfigure[Level 2: L2-Norm of Joint Angle Errors]{
		\centering
		\includegraphics[width=0.45\textwidth]{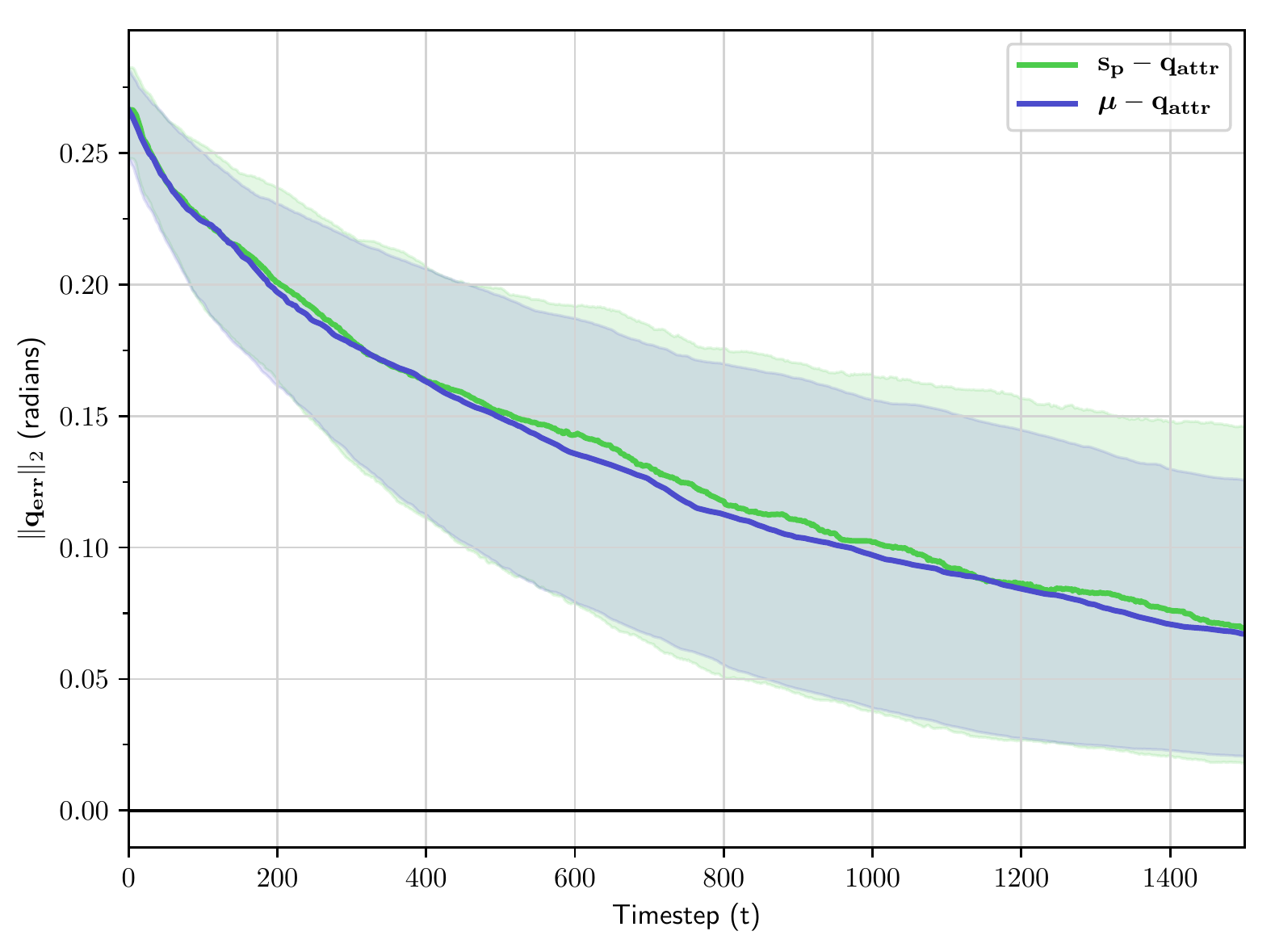}
		\label{fig:muerr_active_level3}
	}	
	\caption{Closer look at simulated NAO active inference results for level 2}
	\label{fig:active_level3_example}
\end{figure*}

\subsection{\textit{Active inference} in the real robot}

\begin{table}[htpb!]
  \caption{Perceptual inference joint angles mean absolute error in the real robot. (in degrees).}
  \label{table:mae_sim1}
  \centering
  \begin{tabular}{lllll}
    \toprule
    Level  & Shoulder Pitch   & Shoulder Roll    & Elbow Yaw        & Elbow Roll        \\  
    \midrule
    1      & 1.326 $\pm$0.823 & 0.679 $\pm$0.773 & 1.569 $\pm$1.441 & 1.9690 $\pm$2.032  \\
    2      & 1.862 $\pm$1.846 & 2.221 $\pm$3.038 & 2.935 $\pm$3.311 & 3.807 $\pm$3.323  \\
    3      & 9.799 $\pm$12.63 & 12.77 $\pm$10.91 & 29.35 $\pm$35.48 & 21.82 $\pm$17.23  \\
    \bottomrule
  \end{tabular}
\end{table}

We tested the proposed algorithm in the real robot. Conversely to simulation, the robot's movements are imprecise due to the mechanical backlash in the actuators ($\pm5^{\circ}$)~\cite{gouaillier2010omni}. Furthermore, we deployed the velocity controller over the built-in NAO position control yielding to a bad synchronization between algorithm and the real movement. For instance, the robot had to wait for the generated actions to be large enough, in order to send the commands to the controller. This caused the movements to be unsmooth. Moreover, as we did not have direct access to the motor driver, the time the action was executed had a large mismatch between the internal error and the actual arm position, resulting in a desynchronization between the internal model and the real world, which could cause the system to diverge.

\begin{figure*}[hbtp!]
	\centering
	\includegraphics[width=0.9\textwidth]{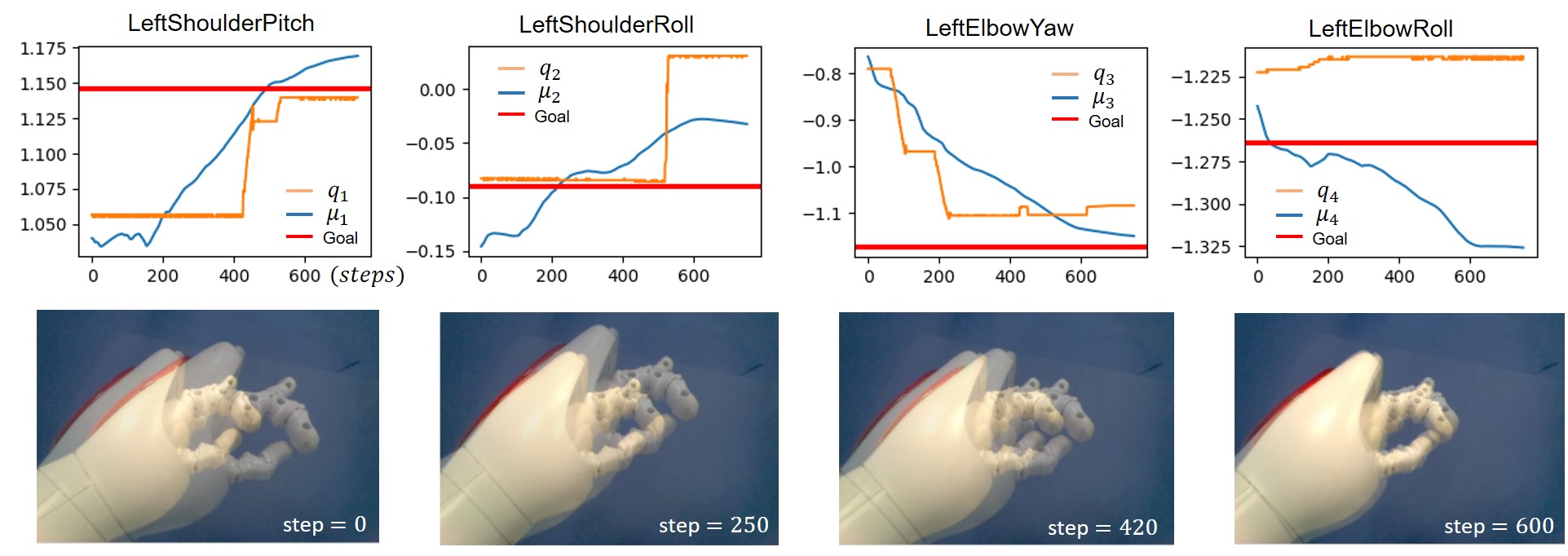}
	\caption{PixelAI test on the real Nao. $\mu$ is the inferred state, $q$ is the real joint angles readings, goal is the ground truth goal angles and $0.05steps = 1 s$. (bottom row) Arm sequence: goal image and Nao visual input are overimposed.}
	\label{fig:pixelAIreal}
\end{figure*}

Furthermore, the visual forward model was expected to model the more complex structure of the real robot hand, that is subject to lighting differences and has a reflective surface. Unlike in the simulator, the same conditions cannot be restored perfectly in the real world, so the model training is always subject to additional noise in the dataset. We used the same deep convolutional decoder architecture for our tests on the real robot as well. Low training error was achieved on the training dataset (MSE in pixel-intensity: ca. 0.0015). The results of the perceptual inference for real NAO on all 3 benchmark levels are shown in Fig.~\ref{fig:all_levels_real}. Similar behaviours of perceptual convergence were found in level 1 and 2, while level 3 had a larger error due to the local minima. Figure \ref{fig:pixelAIreal} shows the PixelAI algorithm running on the robot.

\section{Discussion}
\label{sec:discussion}

We have shown that variational free-energy optimization can work as a general inner mechanism for both perception (estimation) and action (control). Our algorithm extends previous \textit{active inference} works tackling high-dimensional visual inputs and providing sensory generative models learning. This prediction error variant of control as inference \cite{toussaint2009robot} exploits the representation learnt to indirectly generate the actions without a policy. The robot is producing the actions reach the desired goal in the visual space without learning the explicit policy. 

There are other interesting advantages of our proposed approach. First, it is neuroscience-inspired, specifically it grounds on predictive coding approach and the free energy principle~\cite{rao1999predictive,friston2010unified}. This means that we can directly make comparisons with human body perception~\cite{hinz2018drifting}. Secondly, the features learnt instead of being bound to the task, they are grounded to the body. Hence, learnt representations have a physical meaning and can be used to solve other tasks. Furthermore, it does not need rewards engineering and works directly in sensory space. 

A deeper analysis should be performed to evaluate the limits of the algorithm to scale to sequential complex tasks. For instance, we can design a system that generates desired internal beliefs that are transformed into expected sensations that will drive the agent towards the goal, in line with the view of perception as a hierarchical dynamical system~\cite{tani2016exploring}. 



The free energy formulation~\cite{friston2010unified} and variational autoencoders~\cite{kingma2013auto} share the same theory and solve similar problems, as shown in \cite{zambelli2020multimodal}, where the action is directly outputted from the variational autoencoder. However, autoencoders architecture do not account for the original ideas from the Helmholtz machine~\cite{dayan1995helmholtz}. Perception should be an active continuous process~\cite{bajcsy2018revisiting}. Here we have shown how we can use variational inference to provide the active adaptation and interpolation to online input exploiting the forward and backward passes of the neural network. This allows us to incorporate priors (top-down modulation) while maintaining inference dynamics over the observations (bottom-up).

Besides, our PixelAI algorithm complements the active inference community effort to provide scalable models for real applications~\cite{tschantz2019scaling, millidge2019deep}.

\section{Conclusions}
\label{sec:conclusions}
We have described a Pixel-based deep Active Inference algorithm and applied it for robot body perception and action. Our algorithm enabled estimation of the robot arm joint states just using a monocular camera input and perform goal driven behaviours using imaginary goals in the visual space. Statistical results showed convergence in both perception and action in different levels of difficulty with a larger error when dealing with totally random arm poses. This neuroscience-inspired approach is thought to make deeper interpretations than conventional engineering solutions~\cite{hassabis2017neuroscience}, giving some grounding for novel machine learning developments, especially for body perception and action.

\bibliographystyle{plain}
\bibliography{pixel,ninaRHI,pl,selfception,icub}

\end{document}